# Hierarchical method for cataract grading based on retinal images using improved Haar wavelet

Lvchen Cao[1], Huiqi Li[1*], Yanjun Zhang[1], Liang Xu[2], Li Zhang[2]

[1]*School of Information and Electronics, Beijing Institute of Technology, Beijing 100081, China*

[2]*Beijing Institute of Ophthalmology, Beijing Tongren Hospital, Beijing 100730, China*

**Abstract**

Cataracts, which are lenticular opacities that may occur at different lens locations, are the leading cause of visual impairment worldwide. Accurate and timely diagnosis can improve the quality of life of cataract patients. In this paper, a feature extraction-based method for grading cataract severity using retinal images is proposed. To obtain more appropriate features for the automatic grading, the Haar wavelet is improved according to the characteristics of retinal images. Retinal images of non-cataract, as well as mild, moderate, and severe cataracts, are automatically recognized using the improved Haar wavelet. A hierarchical strategy is used to transform the four-class classification problem into three adjacent two-class classification problems. Three sets of two-class classifiers based on a neural network are trained individually and integrated together to establish a complete classification system. The accuracies of the two-class classification (cataract and non-cataract) and four-class classification are 94.83% and 85.98%, respectively. The performance analysis demonstrates that the improved Haar wavelet feature achieves higher accuracy than the original Haar wavelet feature, and the fusion of three sets of two-class classifiers is superior to a simple four-class classifier. The discussion indicates that the retinal image-based method offers significant potential for cataract detection.

## 1. Introduction

A cataract is defined as a lenticular opacity, usually presenting with poor visual acuity. According to a World Health Organization report, the estimated number of people in the world who are blind will exceed 40 million by 2025. More than 50% cases of blindness are caused by cataracts, which are considered as the most common cause of blindness [1–4]. When a patient delays treatment for longer, the visual impairment will be more severe. It is important to improve the quality of eye care service, particularly pre-detection. Although it is widely accepted that early detection and treatment can reduce the suffering of cataract patients, people in less developed regions still cannot receive timely treatment, owing to a lack of professional ophthalmologists [5].

* Corresponding author.
E-mail address: huiqili@bit.edu.cn

Slit lamp imaging and Scheimpflug imaging are commonly used techniques for clinical cataract diagnosis [6]. Cataract diagnosis protocols include the lens opacity classification system (LOCS III) [7], Oxford clinical classification [8], and American cooperative cataract research group (CCRG) method [9]. The LOCS III requires a slit lamp for clinical assessment, while the Wisconsin cataract grading system requires photographic grading, both of which are intricate procedures for most patients and can only be performed by well-experienced ophthalmologists. The Oxford clinical classification and CCRG method exhibit similar problems [7–11].

The ultrasound backscattering signal [12–16] is used for cataract assessment based on the animal model. By using probability density features and multiclass classifiers, the accuracy of cataract hardness assessment achieves 95% when using a small training set. Optical coherence tomography [17] and the ultrasound biomicroscope [18] can provide more accurate screening. However, these imaging techniques are expensive and their operation is complicated, meaning that they are hardly widespread in less developed regions. Thus, it is crucial to simplify the process and reduce costs for early cataract screening.

A simple method for detecting cataracts based on the fundus camera was proposed by the Beijing Tongren Hospital [19–21]. The detection is realized by evaluating the blurriness of retinal images. By taking the diagnosis using a slit lamp as the truth value, the sensitivity when using retinal images achieves 100% (cortical cataract), 84.2% (nuclear cataract), and 76.2% (posterior subcapsular cataract) [21]. The feasibility of this method has been investigated, and inexperienced graders can be trained with several examples (100 image pairs). Compared to ophthalmologists, the accuracy of minimally trained graders reaches 99% [22].

The fundus camera can easily be operated by technologists, or even patients themselves [23]. Since the fundus camera was invented in 1910, retinal images have been used extensively in the diagnosis of ophthalmological diseases, such as glaucoma [24–25], age-related macular degeneration [26–27], and diabetic retinopathy [28–29]. Moreover, this technique has been studied in an attempt to quantify cataract severity automatically [30–31]. Four typical retinal images, representing non-cataract, and mild, moderate, and severe cataracts, are displayed in Figure 1. Figure 1(a) presents a healthy retina, in which the main vessels, optic disc, choroid, and even capillary vessels can be clearly observed. An image with a mild cataract is shown in Figure 1(b), where the main vessels and optic disk are visible, while the choroid and capillary vessels are only faintly visible. In Figure 1(c), only the main blood vessels and optic disc are visible. Furthermore, almost no retinal structures can be observed in Figure 1(d). It can be concluded that less retinal structures can be observed with more severe cataracts.

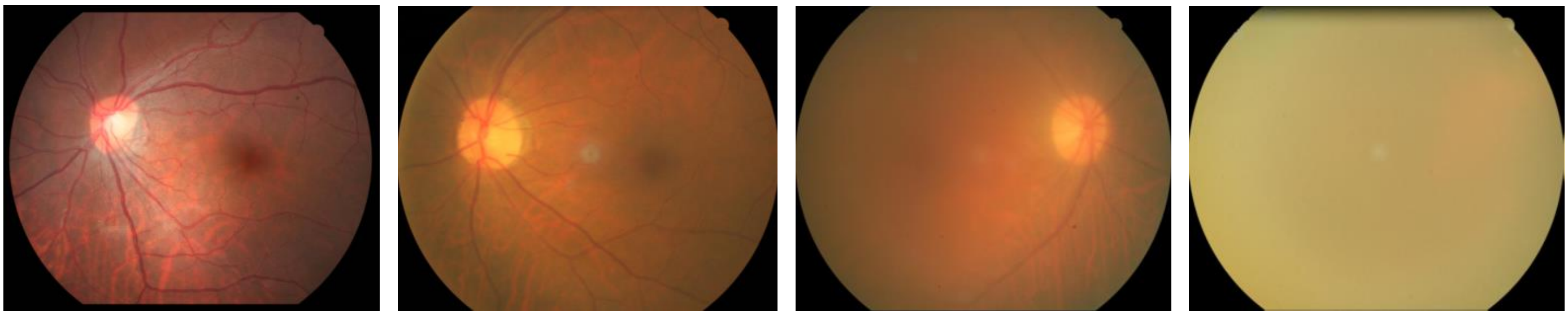

(a) (b) (c) (d)

**Figure 1.** Retinal images: (a) non-cataract, (b) mild cataract, (c) moderate cataract, and (d) severe cataract.

A cataract develops slowly, and may be partial or complete, as well as stationary or progressive. Patients with cataracts of different severities require different treatments. Patients with mild cataracts can be treated by wearing antiglare sunglasses to delay deterioration [32], while surgical assistance is essential for moderate and severe cataract patients [33]. By extracting appropriate features from retinal images, it is possible to distinguish the different levels of severity. This study attempts to develop an automatic method for grading cataract severity based on retinal images. This approach is simple and requires little knowledge, which is meaningful for improving the medical conditions in less developed regions [34].

The main contributions of this paper can be summarized in terms of three aspects. (1) The Haar wavelet transformation is improved based on the retinal image characteristics. By means of this improvement, the contrast of the one-layer detail components is substantially enhanced, and the distortion of the detail components from multilayer decomposition is avoided. (2) Based on the improved Haar wavelet and back-propagation neural network (BP-net), the optimal feature extraction threshold is obtained according to the ergodic searching process. (3) Three sets of two-class classifiers are trained, and a hierarchical classification method is developed on this basis.

The remainder of this paper is organized as follows. Section 2 reviews the related work. Section 3 elaborates on the necessity for improving the Haar wavelet, and details the improvement process. Section 4 presents the methodology applied. Section 5 reports on the performance analysis and provides a discussion. Section 6 presents conclusions and future work.

## 2. Related work

Automatic cataract detection using different imaging modalities has been investigated in recent years [35–42]. Sparse range-constrained learning (SRCL) [35] and the tournament-based ranking convolutional neural network (CNN) [36] have been used to grade the opacity of slit lamp lens images. The SRCL method has achieved higher accuracy than other sparse learning methods, while the tournament-based ranking CNN has effectively solved the problem of unbalanced labels. Posterior subcapsular cataract (PSC) detection based on retro-illumination images using a watershed and Markov random fields method was proposed in [37], and the sensitivity of the PSC screening was 91.2%, based on 519 testing image pairs. Automatic cataract detection using visible wavelength eye images was studied in [38], which is meaningful in developing countries. Moreover, the deep learning scheme has been adopted to detect surgical instruments and surgical phases in cataract videos, which can assist surgeons in standardizing procedures [39].

Automatic cataract detection using fundus images has also attracted significant attention, and was discussed in [40–42]. Prior to presenting the proposed method, the validation of retinal images in cataract screening is discussed and the features are introduced [43–49].

(1) *Validation*: A study for cataract detection based on retinal images using Fourier transform was proposed in [43]. The high-frequency component distributions of retinal images differ owing to their varying levels of blurriness. The distribution curve of the high-frequency component is employed to evaluate the cataract severity. The results demonstrate that the performance of this method exhibits strong correlation with the LOCS III score, which provides the validation of the retinal image-based method.

(2) *Global feature-based method*: The Haar wavelet and discrete cosine transformation (DCT) were used in [44–45], in which four-class classification based on retinal images was studied. The results demonstrated that the Haar wavelet feature achieves higher accuracy than the DCT feature in both the discriminant analysis and supervised method. Texture and morphological features were used in [46], and the results indicated that this approach also achieved relatively high accuracy based on the BP-net. An ensemble learning-based method was proposed in [47], and the performances of the Haar wavelet, DCT, and texture features were summarized. The results demonstrated that the Haar wavelet feature achieves higher accuracy than the DCT and texture features.

(3) *Local feature-based method*: The local standard deviation was used in [48]. The blood vessels were first enhanced by match filtering in different directions, following which the local standard deviations of the coefficients and the other eight features are extracted. Combined with a decision tree, retinal images with different cataract severities were effectively classified.

(4) *Deep feature-based method*: Deep CNN (DCNN)-based classification was applied in [49], and the DCNN achieved slightly higher accuracy compared to the aforementioned methods. However, the accuracy achieves relatively high and stable level when the training samples are close to 2000, thus requiring substantially more images than other related works.

According to the review of the related work, we noted the following points. (1) The scale of the visible retinal structure is the most valuable information for evaluating cataract severity. (2) The most difficult decision-making process in such a classification problem is to distinguish the adjacent two-class retinal images, as they exhibit similar characteristics and it is challenging to capture the difference accurately. (3)

Compared to deep methods, feature extraction approaches can also achieve satisfactory performance, particularly in cases of inadequate training samples. (4) Among the available works, the Haar wavelet exhibits relatively superior performance. (5) The green channel in an RGB image maintains most of the retinal structures, and the following operations in our approach are all performed on the green channel.

# 3. Analysis and improvement of Haar wavelet

In this section, Haar wavelet decomposition based on retinal images is analyzed, following which the improvement of the Haar wavelet is presented. Finally, the performance of the improved Haar wavelet is demonstrated based on retinal images.

## 3.1. Haar wavelet analysis based on retinal images

Wavelet functions have been used extensively in retinal image processing; for example, vessel segmentation [50], health diagnosis [51], retinal recognition [52], and abnormality detection [53]. In recent years, the applicability of this technique has also been proven in cataract detection [54–57]. The Haar wavelet is selected owing to its low computation cost and effective performance in available works.

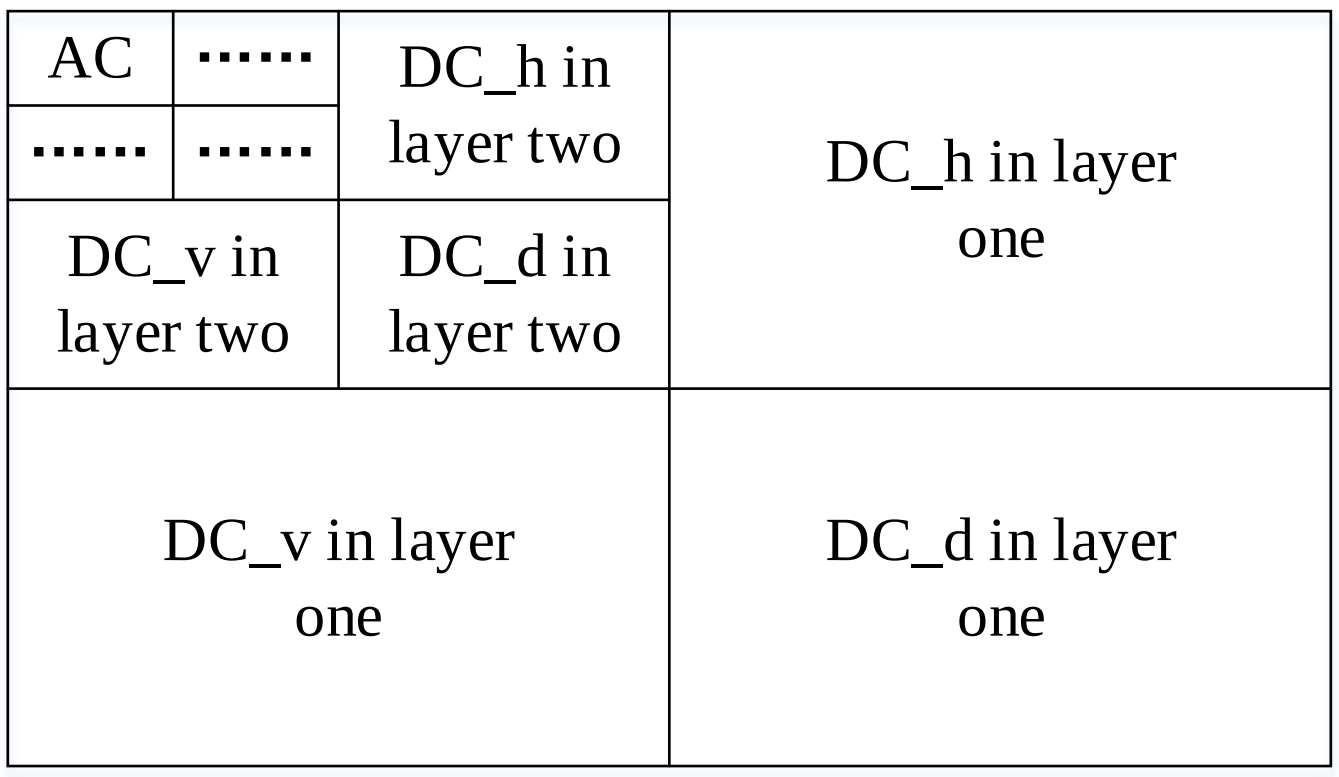

**Figure 2.** Diagram of two-dimensional Haar wavelet decomposition.

Figure 2 illustrates four components generated after a round of Haar wavelet decomposition, including the approximate component (AC) and detail components in the horizontal (DC_h), vertical (DC_v), and diagonal (DC_d) directions, respectively. A further four components can subsequently be obtained by decomposing the AC again. In this manner, multilayer decomposition is achieved. As the intensity difference between the retinal structure and background is small, particularly for blurry retinal images, the texture feature is not obvious following one-layer decomposition. The one-layer detail components of Figures 1(a) to (d) are illustrated in Figures 3(a) to (d). Figures 3(e) to (h) present the corresponding histograms, in which the step is 0.22 and 100 bins are used. It is obvious that the intensity difference of the detail component is tiny among the different images, which makes it difficult to promote effective classification.

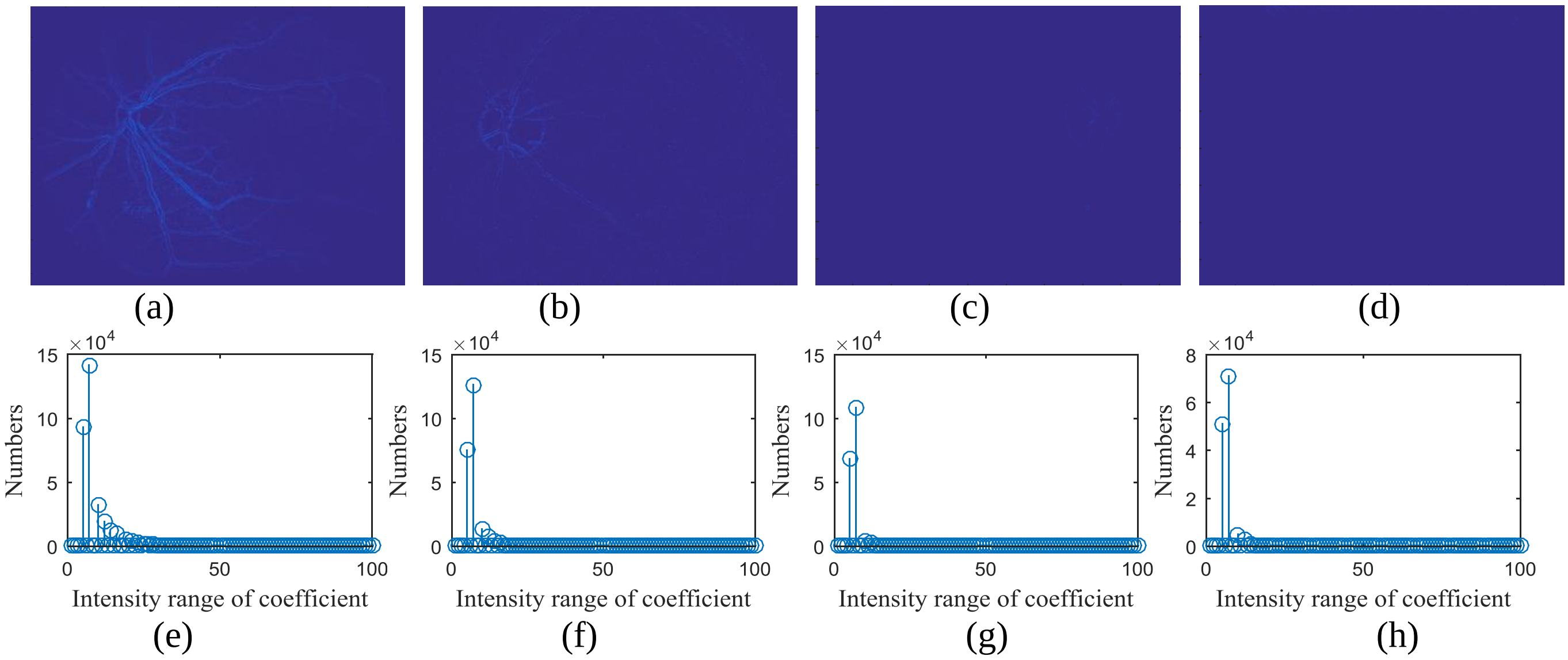

**Figure 3.** Detail components of: (a) non-cataract, (b) mild cataract, (c) moderate cataract, (d) severe cataract, and (e) to (h) histograms of (a) to (d).

Increasing the decomposition layers can effectively enhance the detail component contrast. However, the distortion simultaneously becomes increasingly serious. Figure 1(a) is used as an example to explain this phenomenon, and the detail components from six different layers are illustrated in Figure 4. It can be observed that the vessels and optic disk are not obvious in Figures 4(a) and (b). In contrast, Figures 4(d) to (f) exhibit strong contrast but suffer from serious distortion. The detail component from the third layer achieves a trade-off between the contrast and distortion, which has been employed in Yang's method [45, 47]. However, the size of the third layer detail component is 1/64 compared to the original image, which means that substantial texture features are lost in the multilayer decomposition process. In fact, the detail component of the third layer is generated from the approximate component of the second layer, which has similar visual characteristics but a different size and histogram compared to the original image.

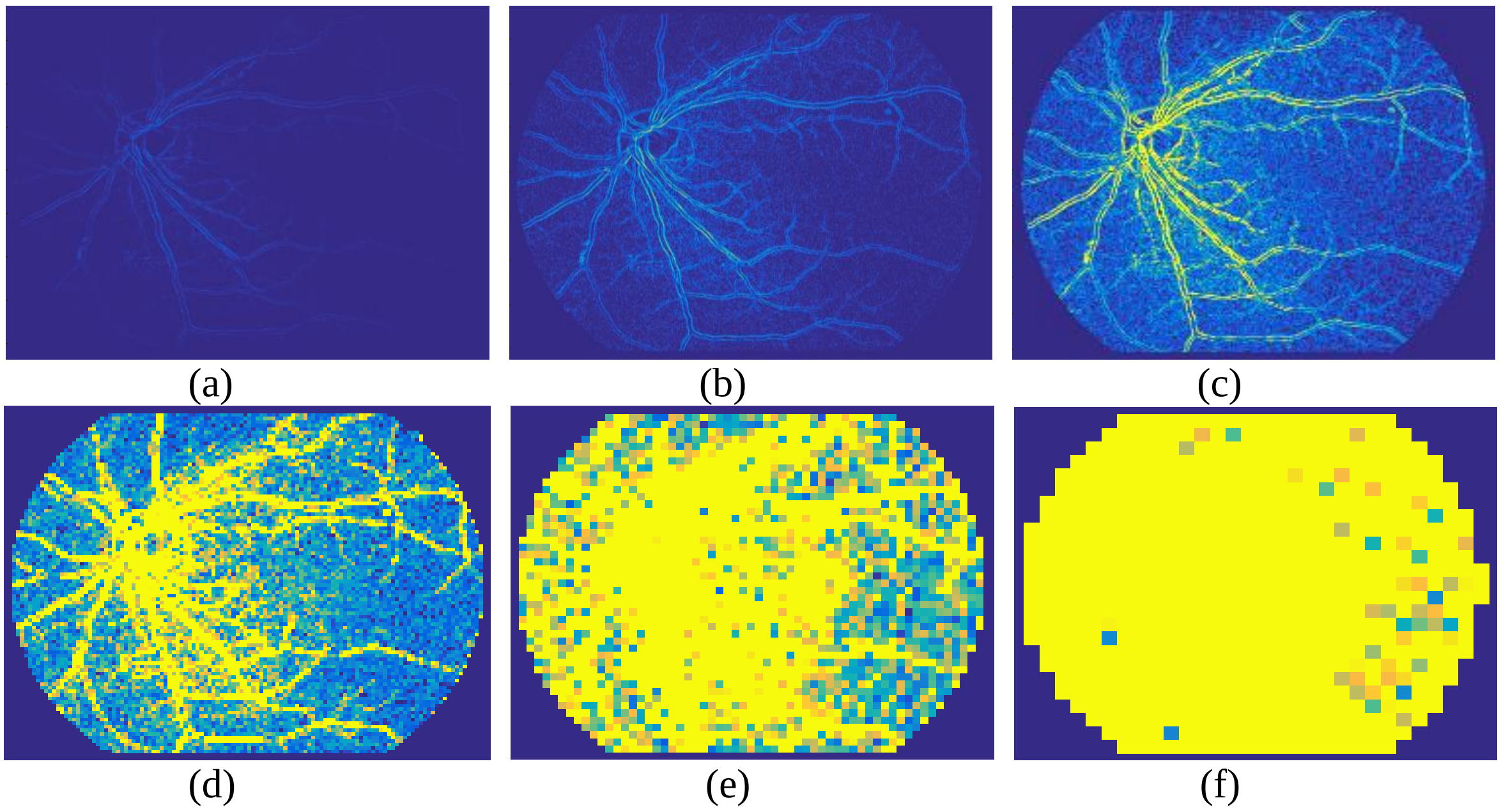

**Figure 4.** Detail components of Figure 1(a): (a) one layer, (b) two layers, (c) three layers, (d) four layers, (e) five layers, and (f) six layers.

It is preferable to use the detail component from one-layer decomposition, because it is directly generated from the original image without distortion. Owing to the one-layer detail component being ineffective for classifying these images, the Haar wavelet decomposition should be improved. The scale and wavelet functions of the Haar wavelet are defined as

$$\begin{cases} \phi(t) = \frac{1}{\sqrt{2}}\phi_{1,0}(t) + \frac{1}{\sqrt{2}}\phi_{1,1}(t), \\ \psi(t) = \frac{1}{\sqrt{2}}\phi_{1,0}(t) - \frac{1}{\sqrt{2}}\phi_{1,1}(t), \end{cases} \tag{1}$$

where $\phi_{j,k}(t)$ is denoted by

$$\phi_{j,k}(t) = \sqrt{2^j}\phi\left(2^j - k\right) \qquad k = 0, 1, \cdots, 2^j - 1, \tag{2}$$

and $\phi_{0,0}(t)$ is defined as

$$\phi_{0,0}(t) = \begin{cases} 0 & t < 0, \\ 1 & 0 \leq t \leq 1, \\ 0 & 1 < t. \end{cases} \tag{3}$$

The decomposition process of a one-dimensional vector {a, b} is illustrated in Figure 5.

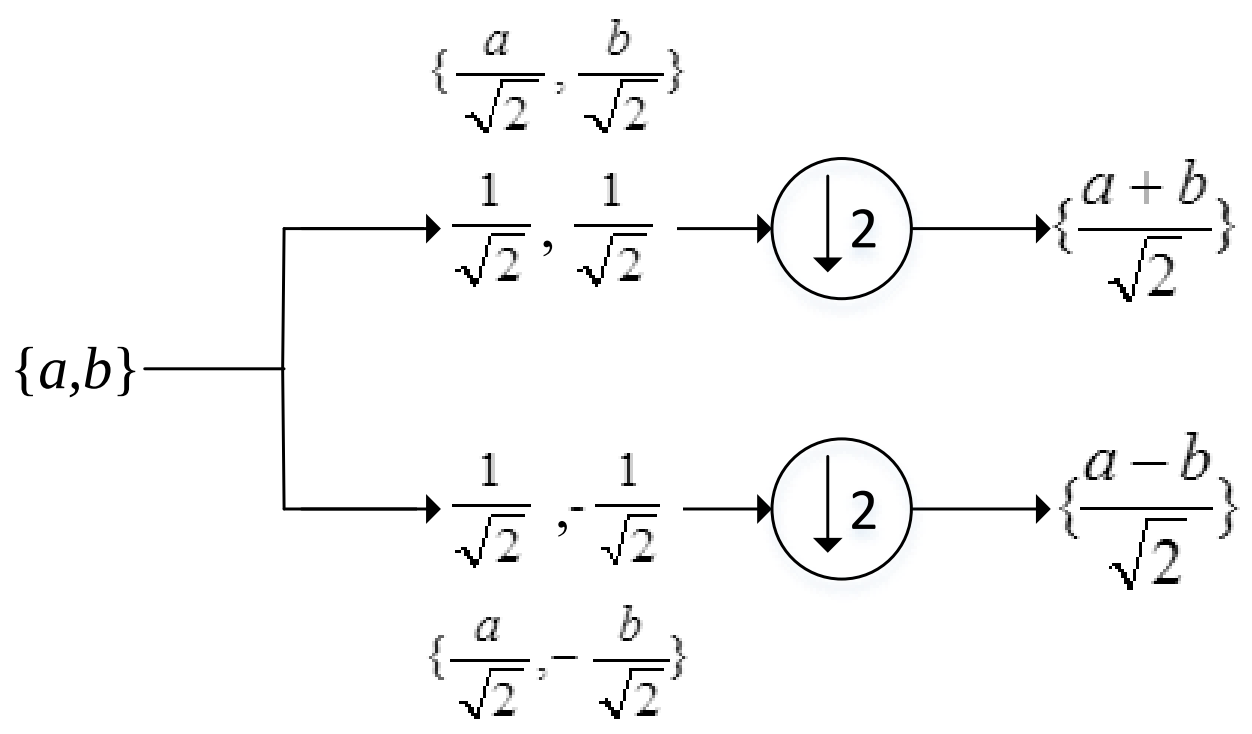

**Figure 5.** Decomposition process of one-dimensional vector.

In Figure 5, $\downarrow 2$ indicates that the step of the down-sampling is 2, while $\{1/\sqrt{2}, 1/\sqrt{2}\}$ and $\{1/\sqrt{2}, -1/\sqrt{2}\}$ are the sampling weights for the approximate and detail components, respectively. Furthermore, $(a+b)/\sqrt{2}$ and $(a-b)/\sqrt{2}$ are the approximate and detail components of $\{a, b\}$, respectively. Therefore, the matrix for calculating the detail component can be expressed as

$$\mathrm{H} = \begin{bmatrix} \frac{1}{\sqrt{2}} & -\frac{1}{\sqrt{2}} & & & & \\ & & \frac{1}{\sqrt{2}} & -\frac{1}{\sqrt{2}} & & \\ & & & \cdots & & \\ & & & & \frac{1}{\sqrt{2}} & -\frac{1}{\sqrt{2}} \end{bmatrix}. \tag{4}$$

The decomposition on a two-dimensional signal (such as a gray image) can be calculated by

$$\mathrm{DC} = \mathrm{H} \cdot \mathrm{I} \cdot \mathrm{H}^T, \tag{5}$$

where I denotes a gray image and $\mathrm{H}^T$ is the transposition of H. Based on (4) and (5), it can be observed that the detail component is calculated from two neighboring pixels. However, the intensity difference between neighboring pixels is often small in blurry retinal images.

### 3.2. Improvement of Haar wavelet

Based on the aforementioned analysis, two limitations should be addressed prior to decomposing the low-contrast images. (1) The detail component from one-layer decomposition is insufficient for highlighting the texture features. (2) Multilayer decomposition produces obvious distortion, which may have a negative effect on the classification.

According to formula (4), the essence of the Haar wavelet decomposition is the convolution operation, in which the kernel is $[1/\sqrt{2}$, -$1/\sqrt{2}]$ and the sliding step is 2. The texture feature extraction can be made more effective by extending the kernel width, because a larger kernel enables the convolution to perceive additional information in the local areas. The extension of the kernel is illustrated as

$$\begin{aligned}
K_1 &= \left[\frac{1}{C_1}, -\frac{1}{C_1}\right] \quad n = 1, \\
K_2 &= \left[\frac{1}{C_2}, \frac{1}{C_2}, -\frac{1}{C_2}, -\frac{1}{C_2}\right] \quad n = 2, \\
K_3 &= \left[\frac{1}{C_3}, \frac{1}{C_3}, \frac{1}{C_3}, -\frac{1}{C_3}, -\frac{1}{C_3}, -\frac{1}{C_3}\right] \quad n = 3, \\
K_n &= \left[\frac{1}{C_n}, \frac{1}{C_n}, \frac{1}{C_n}, \cdots, -\frac{1}{C_n}, -\frac{1}{C_n}, -\frac{1}{C_n}\right] \quad n = n,
\end{aligned} \tag{6}$$

where $K_n$ denotes the improved convolution kernel and $n$ is the half-width of the kernel. When $n$ = 1, formula (6) denotes the original Haar wavelet. Moreover, $C_n$ is the coefficient of the Haar wavelet. According to the orthogonality of wavelet frames [54], the $C_n$ values should satisfy the following formula:

$$\begin{aligned}
&\left(\frac{1}{C_1}\right)^2 + \left(-\frac{1}{C_1}\right)^2 \quad n = 1, \\
&= \left(\frac{1}{C_2}\right)^2 + \left(\frac{1}{C_2}\right)^2 + \left(-\frac{1}{C_2}\right)^2 + \left(-\frac{1}{C_2}\right)^2 \quad n = 2, \\
&= \left(\frac{1}{C_3}\right)^2 + \left(\frac{1}{C_3}\right)^2 + \left(\frac{1}{C_3}\right)^2 + \left(-\frac{1}{C_3}\right)^2 + \left(-\frac{1}{C_3}\right)^2 + \left(-\frac{1}{C_3}\right)^2 \quad n = 3, \\
&= \left(\frac{1}{C_n}\right)^2 + \left(\frac{1}{C_n}\right)^2 + \left(\frac{1}{C_n}\right)^2 + \cdots + \left(-\frac{1}{C_n}\right)^2 + \left(-\frac{1}{C_n}\right)^2 + \left(-\frac{1}{C_n}\right)^2 \quad n = n,
\end{aligned} \tag{7}$$

In formula (7), $\mathrm{C}_1$ is $\sqrt{2}$, and therefore, $\mathrm{C}_2$ and $\mathrm{C}_3$ should be $\sqrt{4}$ and $\sqrt{6}$, respectively. It can be determined that the value of $\mathrm{C}_n$ is $\sqrt{2n}$, according to formula (7). Based on the kernel extension, the texture feature extraction of low-contrast images can be achieved be more easily. The width selection should depend on the characteristics of the specific images. Based on the application in retinal images, the width selection and corresponding performance are provided in the following subsection.

### 3.3. Performance of improved Haar wavelet

Kernel width selection aims to make the texture features of low-contrast images more obvious. According to Figure 1, the texture feature of a small vessel is the most difficult to detect, and its intensity is similar to the surrounding pixels. Therefore, this issue should be considered and the kernel width is set to twice the small vessel width. The width of the narrowest visible vessel is approximately 3 to 4 pixels in our database. To highlight the texture feature fully, $n$ is set to 4, and therefore, $C_n$ is $\sqrt{8}$ according to section 3.2. The new matrix for decomposition can be expressed as

$$\mathrm{H}' = \begin{bmatrix} \frac{1}{\sqrt{8}} & \frac{1}{\sqrt{8}} & \frac{1}{\sqrt{8}} & \frac{1}{\sqrt{8}} & -\frac{1}{\sqrt{8}} & -\frac{1}{\sqrt{8}} & -\frac{1}{\sqrt{8}} & -\frac{1}{\sqrt{8}} & & & & \\ & & \frac{1}{\sqrt{8}} & \frac{1}{\sqrt{8}} & \frac{1}{\sqrt{8}} & \frac{1}{\sqrt{8}} & -\frac{1}{\sqrt{8}} & -\frac{1}{\sqrt{8}} & -\frac{1}{\sqrt{8}} & -\frac{1}{\sqrt{8}} & & \\ & & & & & & & & \cdots & & & \\ & & & & \frac{1}{\sqrt{8}} & \frac{1}{\sqrt{8}} & \frac{1}{\sqrt{8}} & \frac{1}{\sqrt{8}} & -\frac{1}{\sqrt{8}} & -\frac{1}{\sqrt{8}} & -\frac{1}{\sqrt{8}} & -\frac{1}{\sqrt{8}} \end{bmatrix}. \tag{8}$$

Formula (8) means that the eight adjacent pixels are considered as a basic unit; they are multiplied by the corresponding coefficients ($1/\sqrt{8}$ or $-1/\sqrt{8}$) and added together. When this kernel slides through retinal structures, the response will be enlarged.

Figures 6(a) to (c) present three vessel segments with different shapes, intensities, and directions, which are selected to test the improved Haar wavelet performance. The pixel intensities on the three cross-sections are depicted in Figures 6(d) to (f). The corresponding one-layer detail components are illustrated in Figures 6(g) to (i), in which the blue line indicates the results obtained by the original Haar wavelet and the red line represents those of the improved method. It can be observed that the response of the improved method is significantly stronger than that of the original Haar wavelet.

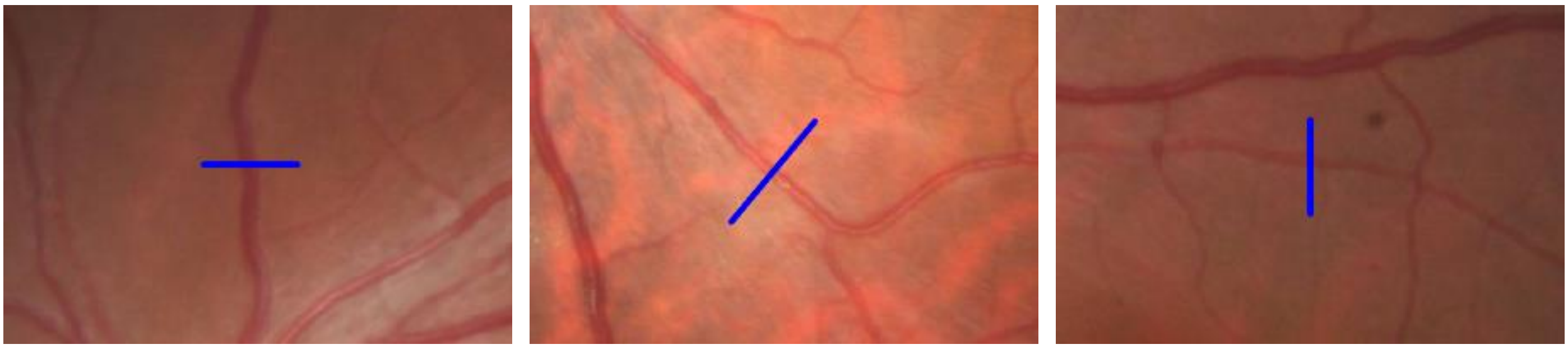

(a) (b) (c)

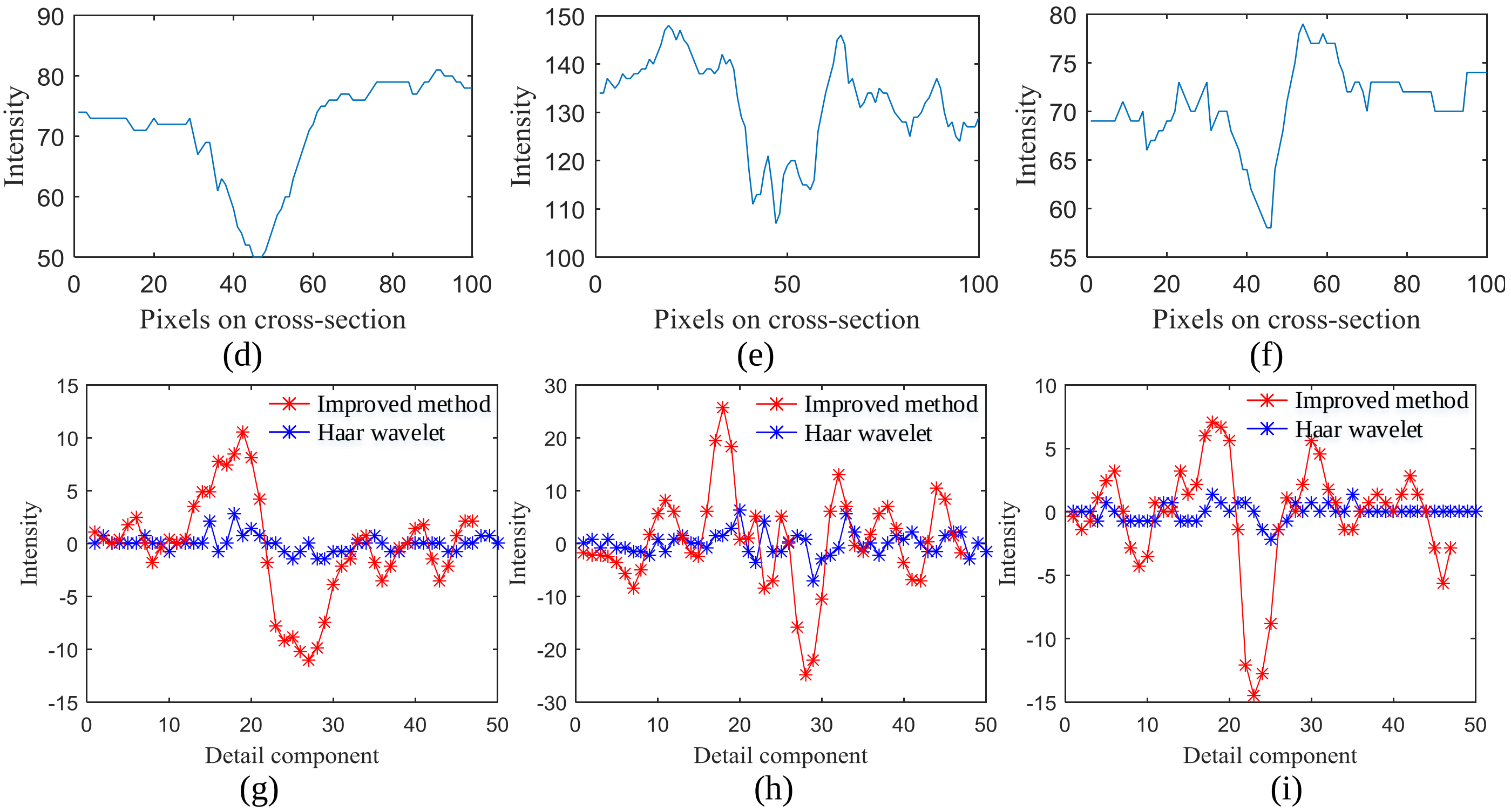

**Figure 6.** Blood vessels in direction of (a) horizontal, (b) diagonal, and (c) vertical; (d) to (f) corresponding intensity distributions on cross-sections; (g) to (i) one-layer detail components based on Haar wavelet and improved method.

Figure 6 indicates that the improved method achieves superior performance on one-dimensional signals. The performance on the complete image is also verified. The detail components of the four retinal images in Figure 1 are presented in Figures 7(a) to (d), in which the blood vessels and optic disk are all highlighted. Compared to Figure 3, the intensity differences among the four images are significantly enlarged. The three-layer detail components obtained by the original Haar wavelet are illustrated in Figures 7(e) to (h), and exhibit similar morphological characteristics compared to the improved method. According to Figure 2, the size of the three-layer detail component is 1/16 of the one-layer component and 1/64 of the original image, which means that the one-layer detail component can represent the original image better. Furthermore, the three-layer detail component suffers from more serious distortion. The performances are compared in the following section.

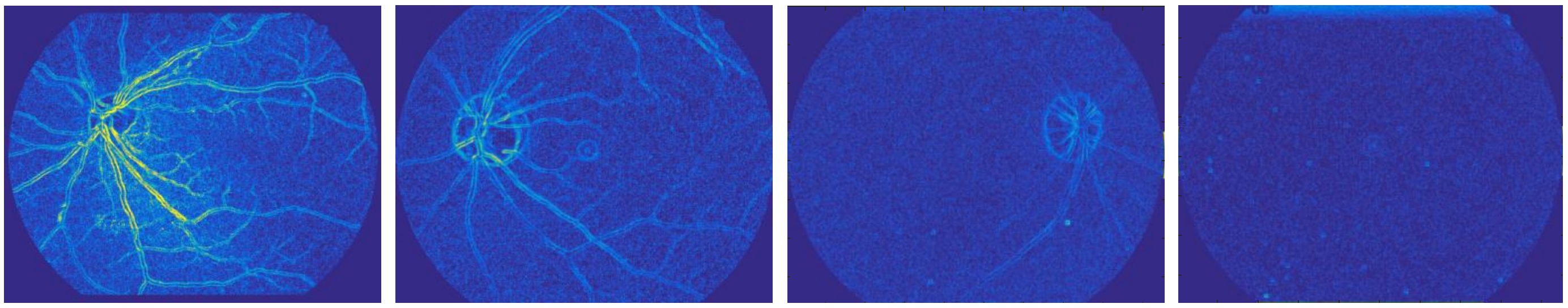

(a) (b) (c) (d)

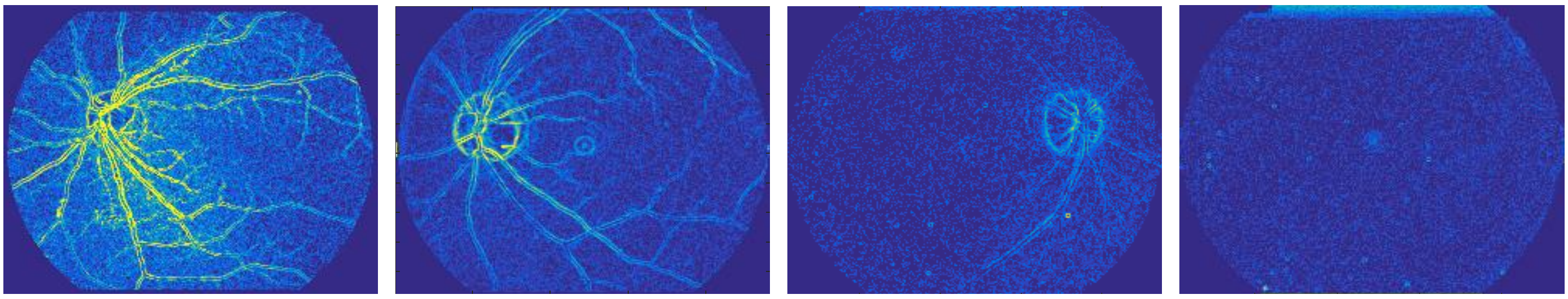

(e) (f) (g) (h)

**Figure 7.** One-layer detail components of (a) non-cataract, (b) mild cataract, (c) moderate cataract, and (d) severe cataract obtained by improved method; three-layer detail components of (e) non-cataract, (f) mild cataract, (g) moderate cataract, and (h) severe cataract obtained by original Haar wavelet.

# 4. Proposed classification strategy

This section presents the pre-processing of the retinal images, feature extraction method, process of searching the optimal threshold for feature extraction, and complete classification method.

## 4.1. Pre-processing of retinal images

Pre-processing is performed to remove the personal information located on the top left corner in order to protect patient privacy. Based on ellipse fitting [55], the retinal boundary is obtained, and the personal information outside the boundary is removed by padding zeros. Thereafter, a certain amount of information residue remains on the retina, as illustrated in Figures 8(a) to (c). A homogenization process is applied to eliminate the patient information completely.

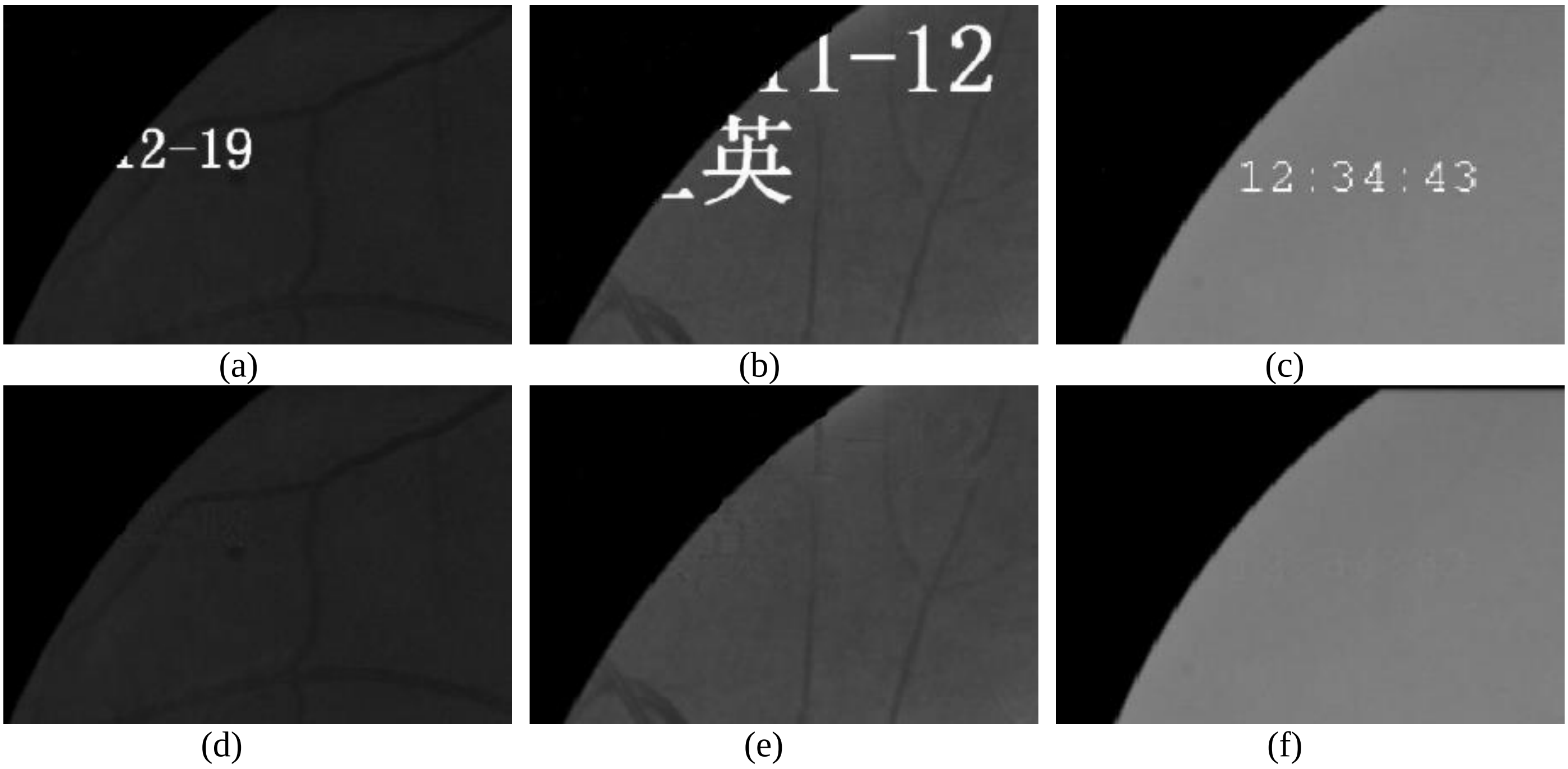

**Figure 8.** (a) to (c) Personal information residual on retinal images; (d) to (f) same screenshots after removing personal information.

The intensity difference between the retinal background and personal information is large, and it is easy to determine the personal information locations according to the threshold. Suppose that the image size is $m \times n$, and the region of $m/4 \times n/2$ on the upper left side is used to label the pixels of personal information. The mean of 100 pixels around the labeled pixels is calculated as M. A random number sequence under the boundary of M − 8 and M + 8 is generated, and the labeled pixels are replaced with the random numbers. According to Figures 8(d) to (f), the personal information is deleted and the edges appear smooth. In this manner, the potential interference factor is reduced. Moreover, a round of median filtering is performed to smooth the original image prior to feature extraction.

Based on the threshold determination and ellipse fitting [55], the retina can be fully segmented. Figure 9 presents the binary image obtained by segmenting Figure 1(a). The maximum width and height are denoted by W and H, and W is always larger than H in our dataset. In this study, feature extraction is based on an annular mask. To adapt the feature extraction method, the image should be resized to guarantee that H is the same as W.

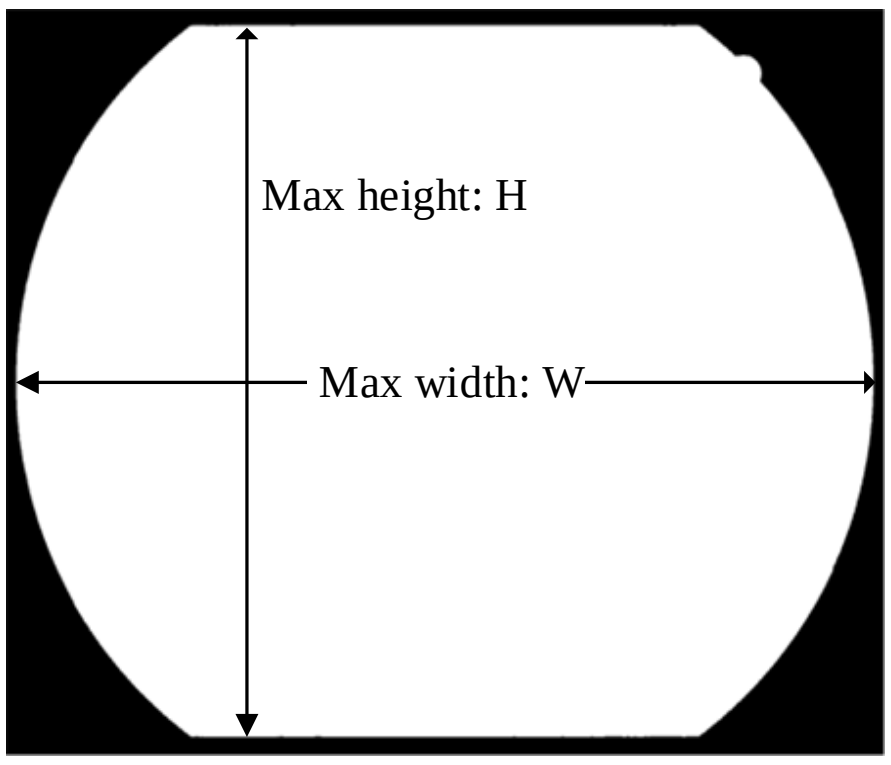

**Figure 9.** Segmentation results of Figure 1(a).

### 4.2. Feature extraction method

The blood vessels are distributed in different directions and the shape of the optic disk is similar to an ellipse. The detail component from one direction is insufficient for presenting the texture feature fully. Moreover, the negative values in the detail component do not mean that they have smaller amplitudes, because they are generated by the opposite gradient direction. Therefore, the absolute values of the detail component in the horizontal, vertical, and diagonal directions are added together in the proposed feature extraction method, and the results are illustrated in Figure 7. We use Figures 7(a) and (b) as examples to present the specific process. Binary images are obtained by segmenting the coefficients in Figures 7(a) and (b) using the same threshold. The results are presented in Figures 10(a) and (b), in which connected regions smaller than 10 pixels are deleted. It is obvious that the non-cataract image yields additional foreground pixels under the same threshold.

Annular masks with different radii are employed in the feature extraction, as illustrated in Figure 10(c). The maximum annular mask radius is N/2, which means that the annular mask should not exceed the boundary of the basic region of interest (ROI) (see Figure 9). All of the annular masks have the same width of (N/2)/20. That is, we use the 20 masks to extract features for each image. The basic ROI is not a standard circle, and cannot be completely covered by the annular masks. Thus, the remaining part and largest annular mask are regarded as one mask. The pixel number in each mask is counted, and the results are plotted in Figure 10(d), in which the smallest mask is labeled as 1 and the largest is labeled as 20. The red and blue lines are generated from Figures 10(a) and (b), respectively. It can be observed that the pixel number in Figure 10(a) is larger than that in Figure 10(b) for each mask.

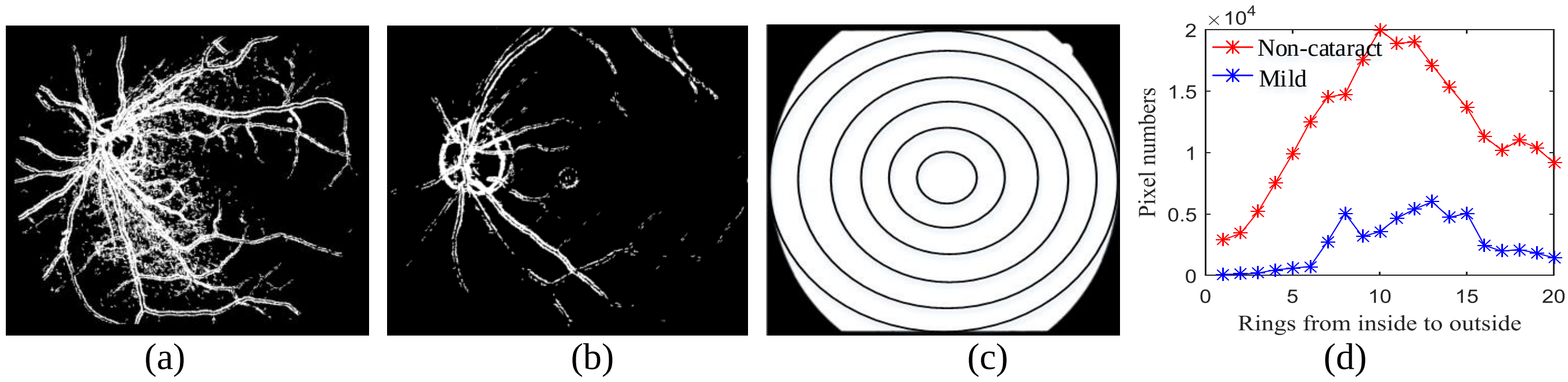

**Figure 10.** (a) and (b) Binary images of Figures 7(a) and (b); (c) sketch map of annular mask; (d) feature curves of (a) and (b).

A total of 10 images from each class are randomly selected, and their feature curves are presented in Figure 11, in which the blue, red, black, and cyan lines represent the non-cataract, mild cataract, moderate cataract, and severe cataract, respectively. Each image contains 20 feature curves: Figure 11(a) depicts 10 images with non-cataract and 10 images with mild cataracts, Figure 11(b) depicts 10 images with mild cataracts and 10 images with moderate cataracts, and Figure 11(c) depicts 10 images with moderate cataracts and 10 images with severe cataracts. These feature curves with different colors exhibit different distributions, and the distribution overlay is minimal. This phenomenon proves that this feature extraction method offers strong potential for classifying retinal images with different severities.

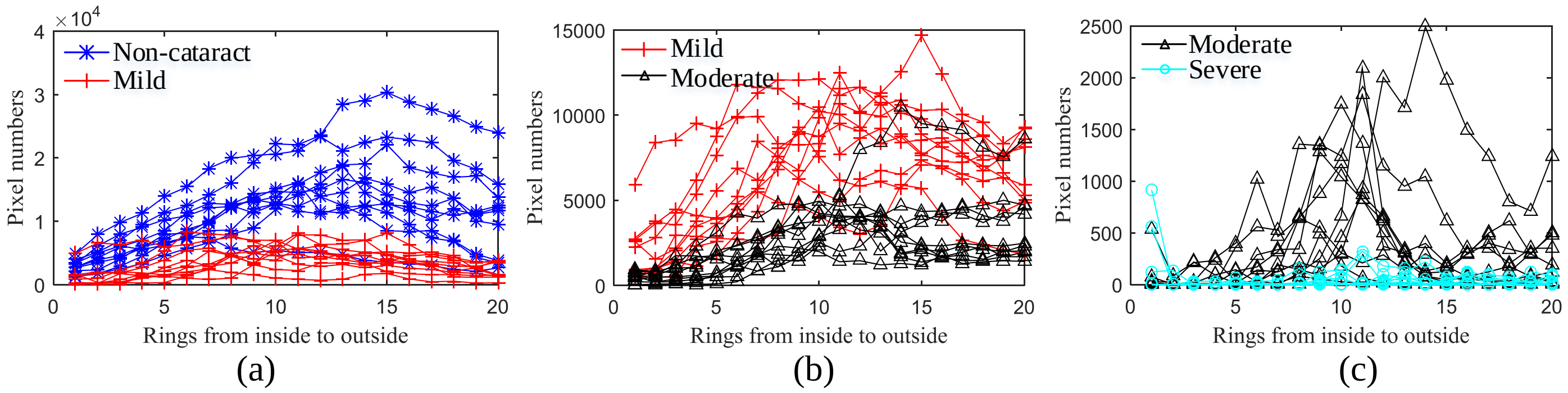

**Figure 11.** Feature curves of (a) non-cataract and mild cataract, (b) mild and moderate cataracts, and (c) moderate and severe cataracts.

### 4.3. Process of searching optimal threshold

The threshold value is set to segment the wavelet coefficients. Coefficients with a higher intensity than the threshold are retained; otherwise, they are discarded. Thereafter, a binary image is obtained for the feature extraction process. A higher threshold means that less foreground pixels are obtained. The three thresholds in Figures 11(a), (b), and (c) are set to 14, 10, and 11, respectively. The three thresholds are learned from the retinal images, and can be considered as relatively effective for the segmentation. For example, when the threshold is 3, the binary images of Figures 7(a) and (b) are illustrated in Figures 12(a) and (b). The corresponding feature curves are presented in Figure 12(c). It is difficult to determine which images exhibit cataracts from either the binary image or feature curves. Therefore, it is essential to establish the optimal threshold in order to segment the detail component between the non-cataract and mild cataract, mild and moderate cataracts, and moderate and severe cataracts, respectively.

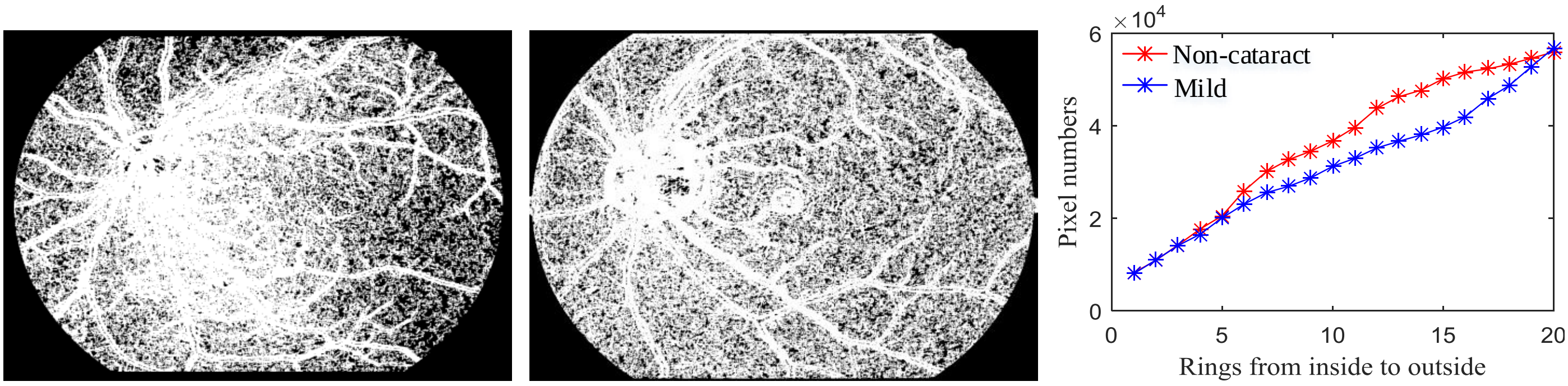

(a) (b) (c)

**Figure 12.** (a) and (b) Binary images of Figures 7(a) and (b) under threshold value of 3; (c) corresponding feature curves of (a) and (b).

The process of searching for the optimal threshold value is based on the loss function of the BP-net, namely the mean square error (MSE). A smaller MSE value often implies higher accuracy in classification tasks. In the proposed method, the thresholds for each two-class classifier are selected when the MSEs are the three smallest values. The interval for the searching process is (1, 40), because this interval contains almost all of the wavelet coefficients. We use the images with non-cataract and mild cataract as an example to demonstrate the process. In this case, 100 non-cataract images and 100 mild cataract images are randomly selected for the searching process. With a step of 1, 40 thresholds are used and 40 feature curve sets are generated. Then, the 40 feature sets are sent into the BP-nets and 40 MSE values are obtained from each training instance.

In our method, the length of the feature vector is 20, which is not very complicated. Thus, a single hidden-layer neural network is used for classification. The formula $(\sqrt{N+M}+a)$ is widely accepted for selecting the number of neurons in the hidden layer [56], where N and M denote the input and output dimensions, respectively, and $a$ is an integer ranging from 0 to 10. Accordingly, the number of hidden layer neurons should be (5 ~ 15) in our method. A lower number of neurons may lead to significant training and generalization errors owing to the under-fitting problem, while excessive neurons results in low training errors and high generalization errors owing to over-fitting [57]. The value is set to 10 in our method, which provides a trade-off between under-fitting and over-fitting.

As illustrated in Figure 13(a), the MSE curve exhibits the smallest value when the threshold is 14. In the same manner, the MSE curves of the mild and moderate cataracts, and moderate and severe cataracts are presented in Figures 13(b) and (c), respectively, in which the two smallest MSE values correspond to thresholds of 10 and 11, respectively.

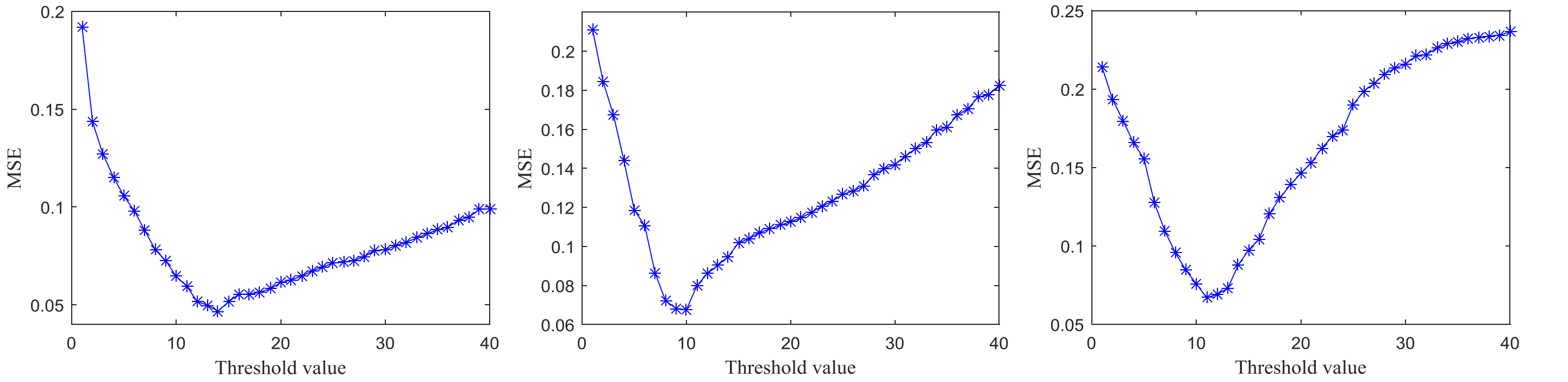

(a) (b) (c)

**Figure 13.** MSE curves by training (a) non-cataract and mild cataract, (b) mild and moderate cataracts, and (c) moderate and severe cataracts.

The receiver operating characteristic (ROC) curves of the three two-class classifiers under optimal threshold values (14, 10, and 11) are presented in Figure 14. The proportion of the training and testing sets is 2:1. In Figure 14, FPR denotes the false positive rate, while TPR denotes the true positive rate. The areas under the three ROC curves are 0.91, 0.88, and 0.84, respectively.

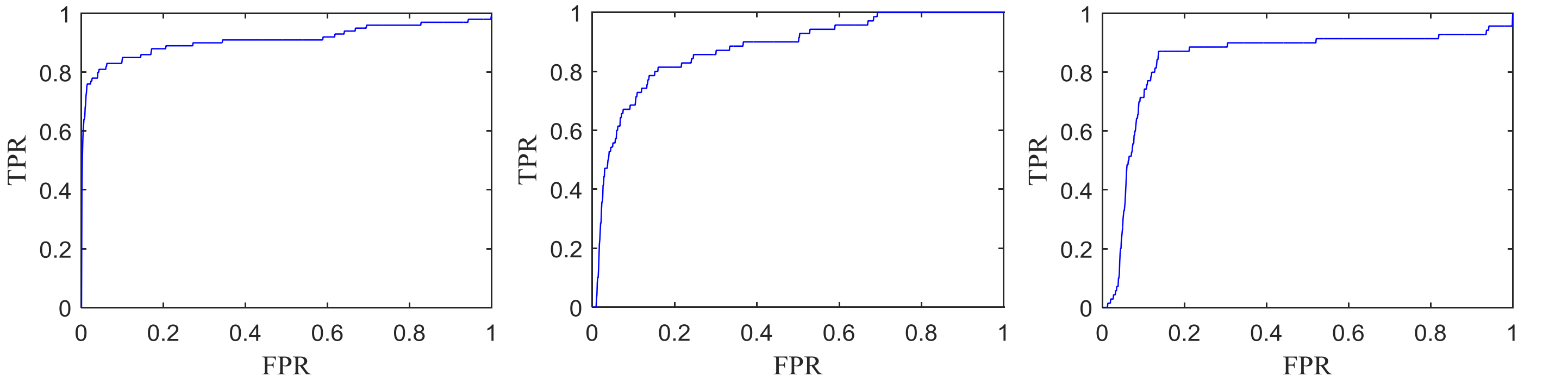

(a) (b) (c)

**Figure 14.** ROC curves based on (a) non-cataract and mild cataract, (b) mild and moderate cataracts, and (c) moderate and severe cataracts.

### 4.4. Complete classification method

The complete classification method is based on majority voting, in which the two-class classifiers based on the three smallest MSE values are selected for the voting. A flowchart of the classification method is presented in Figure 15. Following pre-processing, three feature sets are first extracted based on thresholds of 13, 14, and 15. Then, these are sent into the corresponding classifiers, namely C1 to C3, which have already been trained. Thereafter, majority voting is applied to obtain the final decision-making. In the flowchart, non-cataract is labeled as 1 and the other images are all labeled as 2 in the first classification. In the same manner, the mild images are selected in the second classification and labeled as 2. The moderate images are selected in the third classification and labeled as 3. Finally, the remaining images are considered as severe cataracts and labeled as 4. Following classification, each patient should be clinically treated according to the different severities; for example, severe cataract patients require the earliest surgical intervention.

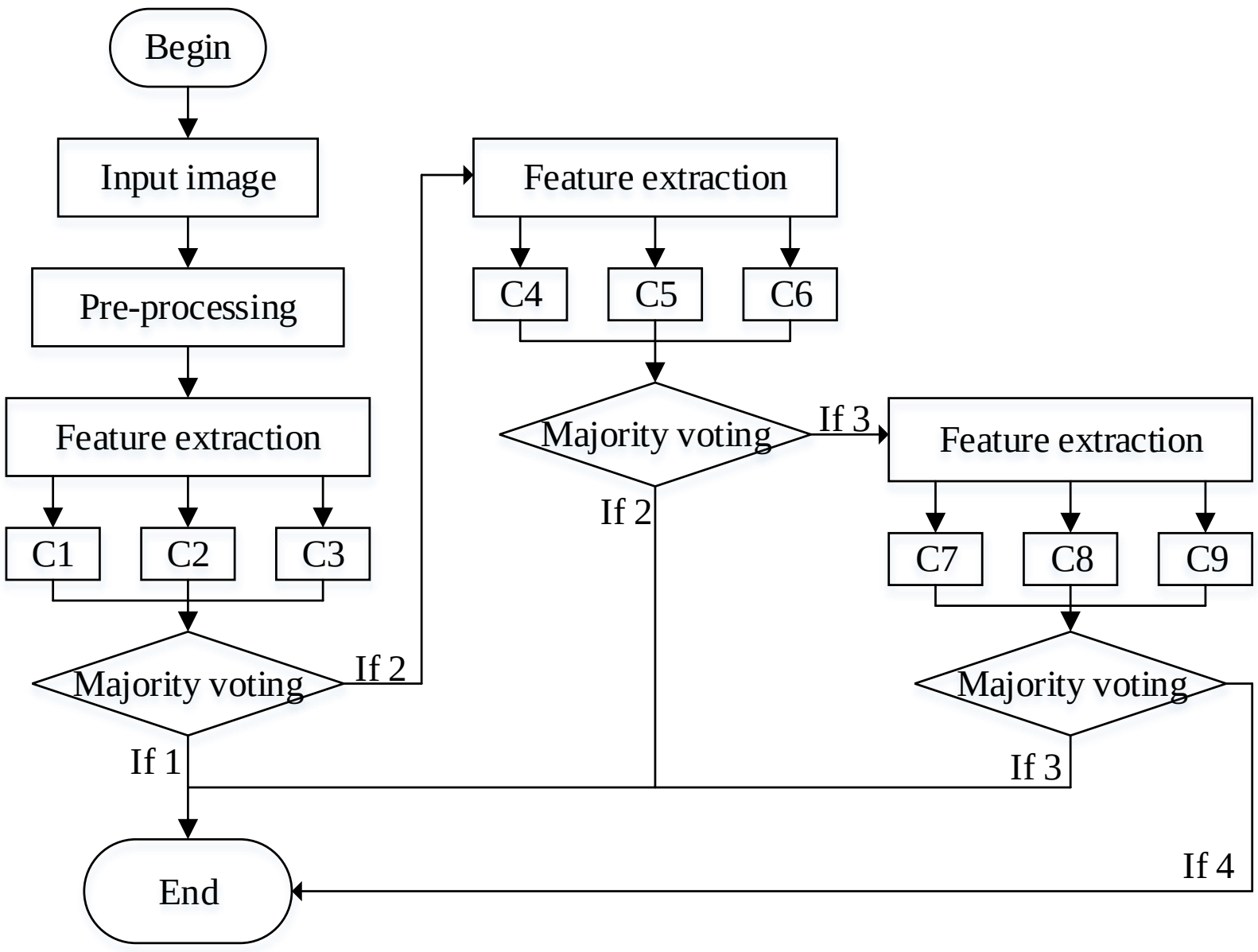

**Figure 15.** Flowchart of complete classification method.

## 5. Results and performance analysis

Among the 1355 retinal images used, 922 were collected from consecutive cataract cases and 433 were non-cataract cases from the Beijing Tongren Hospital. A Canon CR-2 AF non-mydriatic fundus camera was used to capture the retinal images, in which 1340 images were centered on the macula and 15 were centered on the optic disc, all taken at a 45° view angle. These retinal images were graded by two experienced ophthalmologists, according to the criterion described in section 1. In the retinal images, 433 eyes were confirmed as non-cataract, while 415, 217, and 290 images were diagnosed as mild, moderate, and severe cataracts, respectively.

These images were divided into three groups. When one group was used as the testing set, the other two groups were the training set, which ensured that no image appeared in both the training and testing sets. The measures of sensitivity (SE), specificity (SP), and accuracy (ACC) were employed to evaluate the performance. SE and SP represent the ratios of correctly classified target and non-target images. In the two-class classification, retinal images with cataracts were considered as target images. The non-cataract, and mild, moderate, and severe cataracts were considered as the target images in the four-class classification problem, respectively. The results of the two-class and four-class classification are presented in Tables I and II, respectively.

**Table I.** Results of two-class classification

| | Cataract (SE%) | Non-cataract (SP %) | ACC |
|---|---|---|---|
| Group 1 | 95.75 | 95.83 | 95.78 |
| Group 2 | 94.77 | 93.06 | 94.22 |
| Group 3 | 95.81 | 91.72 | 94.51 |
| Total | 95.44 | 93.53 | 94.83 |

**Table II.** Results of four-class classification

| Group | Non-cataract | | Mild | | Moderate | | Severe | | ACC |
|---|---|---|---|---|---|---|---|---|---|
| | (SE %) | (SP %) | (SE %) | (SP %) | (SE %) | (SP %) | (SE %) | (SP %) | |
| Group 1 | 95.83 | 95.75 | 86.23 | 93.91 | 70.83 | 94.71 | 88.54 | 97.74 | 87.33 |
| Group 2 | 93.06 | 94.77 | 84.06 | 94.55 | 70.83 | 94.97 | 89.58 | 97.18 | 86.00 |
| Group 3 | 91.72 | 95.81 | 76.26 | 93.06 | 78.08 | 93.99 | 90.82 | 97.20 | 84.62 |
| Total | 93.53 | 95.44 | 82.17 | 93.83 | 73.27 | 94.55 | 89.66 | 97.37 | 85.98 |

The performance analysis is carried out in the following subsections. The main analysis includes three aspects: (1) the kappa coefficient; (2) a performance comparison with a simple four-class classifier based on the same feature; and (3) a performance comparison with the same classification method based on the original Haar wavelet feature. Moreover, the robustness to noise conditions, a comparison with related works, and a discussion are presented.

### 5.1. Kappa coefficient

The kappa coefficient provides a measure of diagnostic reliability [58]. It is applied extensively in clinical studies and skill assessments [59–60]. The kappa coefficient is based on the confusion matrix, as indicated in Table III, in which HE denotes the classification by human experts in this field, and PR is the predicted result. The formula for computing the kappa coefficient is expressed as

$$Kappa = \frac{p_o - p_e}{1 - p_e}, \tag{9}$$

where $p_o$ and $p_e$ are denoted by

$$p_o = \frac{\sum_{i=1}^{i=n} a_{i,i}}{\sum_{i,j=1}^{i,j=n} a_{i,j}}, \tag{10}$$

$$p_e = \frac{\sum_{j=1}^{j=n} a_{1,j} \times \sum_{i=1}^{i=n} a_{i,1} + \sum_{j=1}^{j=n} a_{2,j} \times \sum_{i=1}^{i=n} a_{i,2} + \cdots\cdots + \sum_{j=1}^{j=n} a_{n,j} \times \sum_{i=1}^{i=n} a_{i,n}}{\sum_{i,j=1}^{i,j=n} a_{i,j} \times \sum_{i,j=1}^{i,j=n} a_{i,j}}. \tag{11}$$

**Table III.** Diagram of confusion matrix

| HE / PR | Category 1 | Category 2 | …… | Category n |
|---|---|---|---|---|
| Category 1 | $a_{11}$ | $a_{12}$ | | $a_{1n}$ |
| Category 2 | $a_{21}$ | $a_{22}$ | | $a_{2n}$ |
| …… | | | …… | |
| Category n | $a_{n1}$ | $a_{n2}$ | | $a_{nn}$ |

The range of the kappa coefficient is (− 1, 1), but it commonly falls into (0, 1). A previous study demonstrated that a diagnosis result is reliable if the kappa value is greater than 0.6 [61], and a higher value is preferable. According to formulae (9) and (10), the kappa coefficient of our strategy is 0.81, which proves that our method exhibits relatively high reliability.

### 5.2. Comparison with simple four-class classifier

In the proposed method, the non-cataract images are first recognized, following which the mild and moderate cataract images are identified successively, and the remaining images are considered as severe cataracts. In this manner, a four-class classification problem is transformed into three two-class classification problems. For a simple four-class classier, the number of output layer neurons is 4, and the other settings remain the same as in the proposed method. The MSE distribution using the four-class classifier is illustrated in Figure 16, where the MSE curve exhibits the three smallest values when the threshold is 9, 10, and 11. Furthermore, majority voting is used, and the SE values of the three groups are presented in Figure 17. The blue line is generated by the simple four-class classifier, while the red line is generated by the proposed method. The non-cataract, and mild, moderate, and severe cataracts are represented by 1, 2, 3, and 4, respectively. It is obvious that the proposed method provides superior performance to a simple four-class classifier.

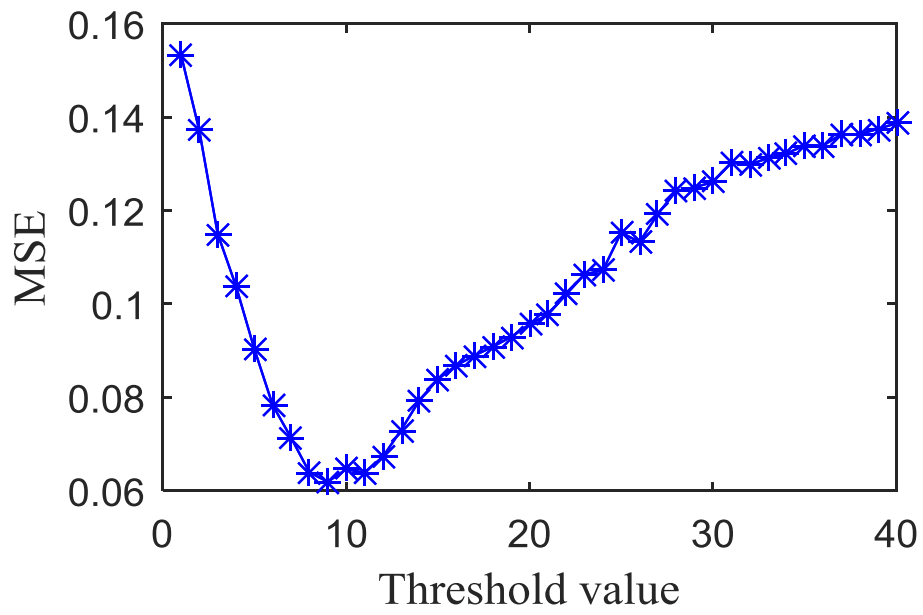

**Figure 16.** MSE distribution of simple four-class classifier.

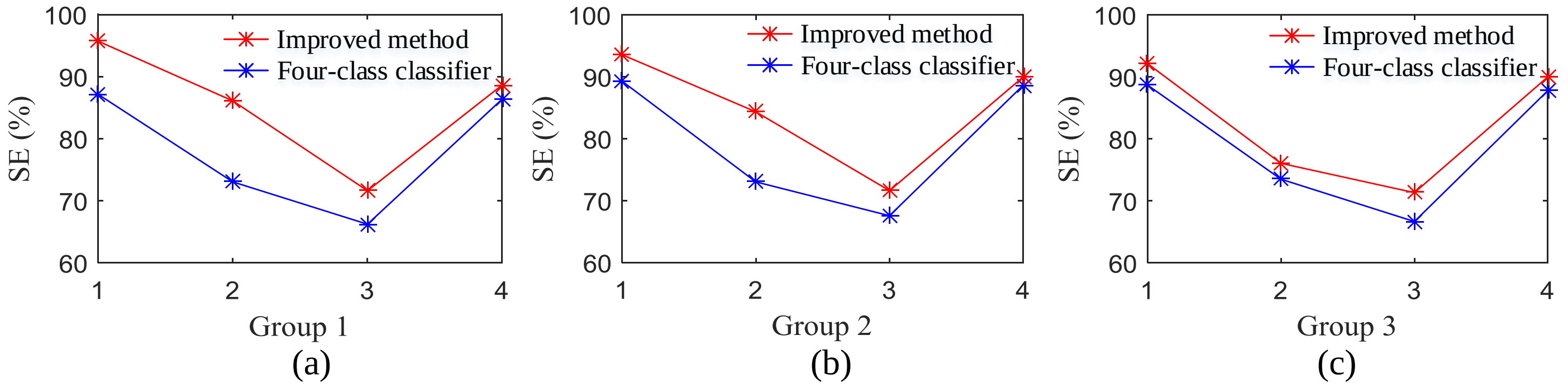

**Figure 17.** Comparison of proposed method and simple four-class classifier: (a) SE of group 1, (b) SE of group 2, and (c) SE of group 3.

### 5.3. Comparison with original Haar wavelet-based features

According to Figure 7, the improved Haar wavelet can obtain proper detail components with only one-layer decomposition, while three-layer decomposition is required by the original Haar wavelet. Although the contrast is enhanced based on the multilayer decomposition, the detail component size is reduced and distortion may exist. Using the same settings as in the proposed method, the MSE curves for the original Haar wavelet for each two-class classifier are presented in Figure 18. The blue line is generated by the original Haar wavelet, while the red line is based on the improved method. The improved method achieves smaller MSE values in all three two-class classifiers, which implies that it can achieve superior classification performance. In Figure 19, the blue line denotes the SE of the non-cataract, and mild, moderate, and severe cataracts based on the original Haar wavelet, while the red line denotes those of the improved method. The images with non-cataract, and mild, moderate, and severe cataracts are denoted by 1, 2, 3, and 4, respectively. It can be observed that the improved method exhibits superior performance to the original Haar wavelet-based method.

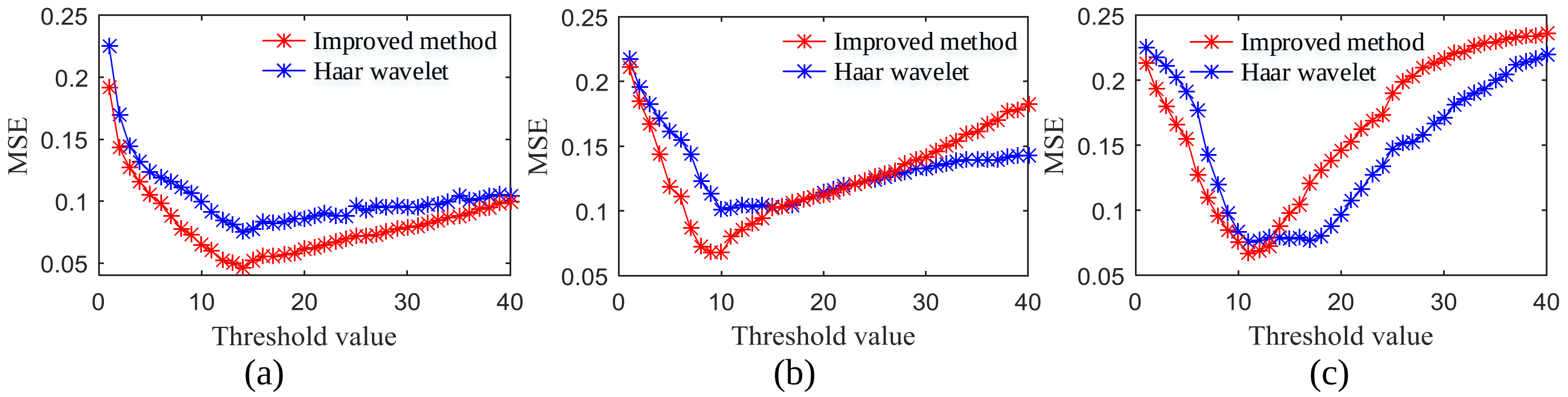

**Figure 18.** MSE curves of (a) non-cataract and mild cataract, (b) mild and moderate cataracts, and (c) moderate and severe cataracts.

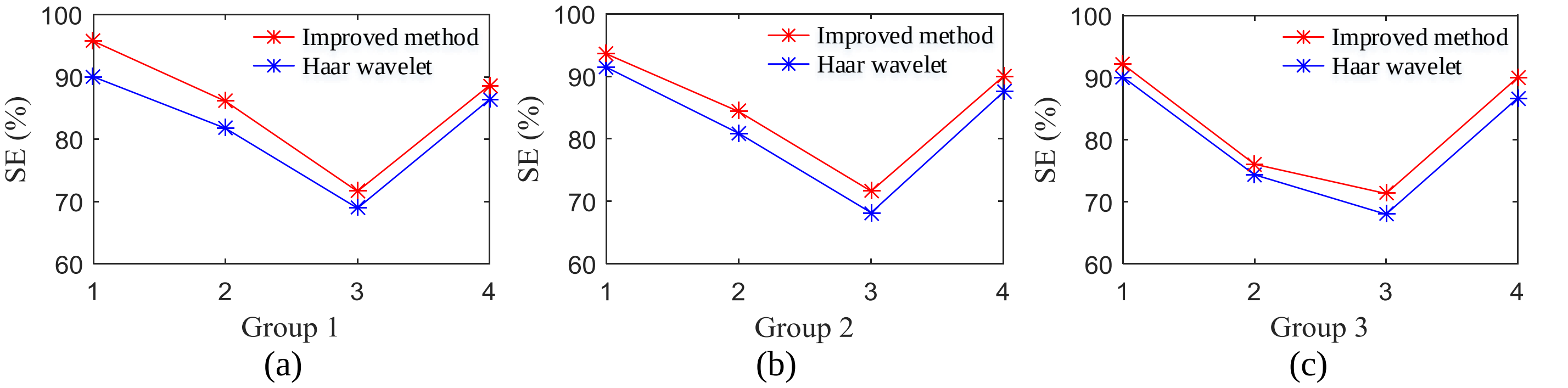

**Figure 19.** Comparison of improved method and original Haar wavelet: (a) SE of group 1, (b) SE of group 2, and (c) SE of group 3.

### 5.4. Performance under noise conditions

Retinal images may be unusable for retinal evaluation when adding moderate noise (for example, 5% salt and pepper noise) [62]. Thus, the retinal images are intentionally corrupted with noise to evaluate the proposed method robustness. Three different noise types are added. For example, the salt and pepper noise-corrupted images are illustrated in Figure 20. According to a study on the effects of noise, the suggested intensity when using Gaussian and speckle noise should be 0 mean and $10^{-3}$ standard deviation [63–64], and salt and pepper noise should affect 5% of the retinal images [65]. The corrupted images are retrained and tested, and the results are detailed in Table IV. The SE declines slightly after adding Gaussian and salt and pepper noise, while the effect of adding speckle noise is negligible. The potential reasons are explained in terms of two aspects: (1) the median filter is operated prior to feature extraction on both the training and testing sets; and (2) connected regions of small sizes are deleted after segmenting the wavelet coefficients. Therefore, the proposed method exhibits relatively strong robustness to noise conditions.

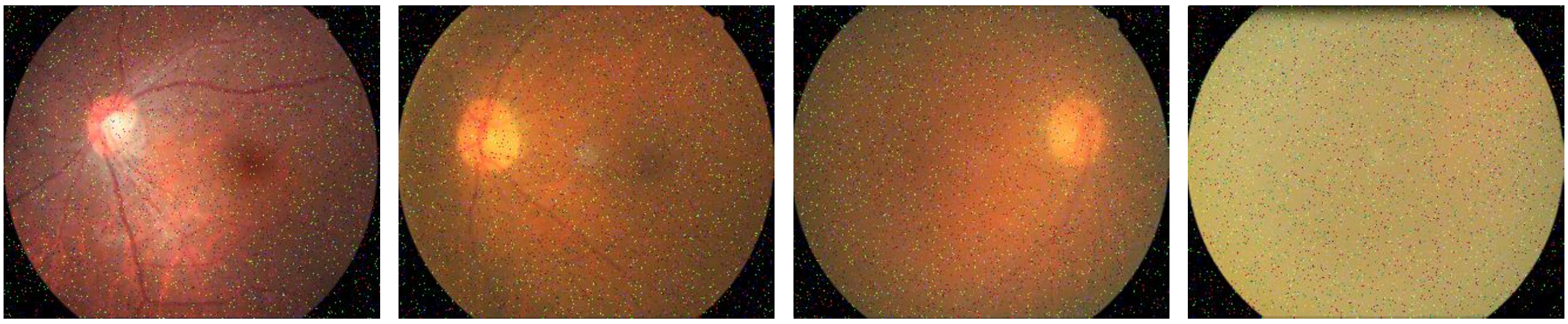

**Figure 20.** Images corrupted by salt and pepper noise.

**Table IV.** SE after adding different noise

| Type of noise | Non-cataract (SE %) | Mild (SE %) | Moderate (SE %) | Severe (SE %) |
|---|---|---|---|---|
| Noise free | 93.53 | 82.17 | 73.27 | 89.66 |
| Gaussian | 87.99 | 78.07 | 69.59 | 88.28 |
| Salt and pepper | 91.69 | 80.72 | 70.05 | 88.62 |
| Speckle | 92.61 | 81.20 | 72.81 | 89.31 |

### 5.5. Summary and discussion

To the best of the authors' knowledge, the results of available works are summarized in Table V. The accuracy is higher than 89% (two-class) and 73.8% (four-class) in all available works. This proves that the retinal image-based method offers strong potential for automatically detecting and grading cataracts.

As the available works are not open source and certain essential settings cannot be found in their publications, only the methods of [43] and [46] are implemented. Ref. [48] is our former work. The performances on our database are presented in Table VI. It can be observed that our method exhibits higher accuracy in both the two-class and four-class classification tasks.

**Table V. Summary of available works**

| Method | | Two-class (unit: %) | | | Four-class (unit: %) | | | | | No. of images |
|---|---|---|---|---|---|---|---|---|---|---|
| | | Cataract (SE) | Non-cataract (SP) | ACC | Non-cataract (SE) | Mild (SE) | Moderate (SE) | Severe (SE) | ACC | |
| Ref. [43] | | N/A | N/A | N/A | N/A | N/A | N/A | N/A | N/A | 330 |
| Ref. [44] | | N/A | N/A | N/A | N/A | N/A | N/A | N/A | 84.77 | 445 |
| Ref. [45] | | N/A | N/A | 89.3 | N/A | N/A | N/A | N/A | 73.8 | 445 |
| Ref. [46] | | N/A | N/A | 89.47 | N/A | N/A | N/A | N/A | 82.9 | 504 |
| Ref. [47] | Voting | 94.2* | 91.5 | 93.2 | 91.00 | 76.5 | 77.8* | 62.5 | 83.9 | 1239 |
| | Staking | 91.2 | 92.5 | 92 | 89.3 | 79.5 | 74.6 | 75.0 | 84.5 | |
| Ref. [48] | | 93.06 | 92.15 | 92.77 | 92.15 | 75.66 | 74.11 | 95.52* | 83.84 | 1355 |
| Ref. [49] | | N/A | N/A | N/A | N/A | N/A | N/A | N/A | 82.51 | 401 |
| | | N/A | N/A | N/A | N/A | N/A | N/A | N/A | 84.64 | 802 |
| | | 92.53 | 94.84* | 93.52* | 95.63* | 83.28* | 57.92 | 81.67 | 86.69* | 5620* |

∗The value is maximum in the corresponding column.

**Table VI. Comparison with related works**

| Method | Two-class (unit: %) | | | Four-class (unit: %) | | | | |
|---|---|---|---|---|---|---|---|---|
| | Cataract (SE) | Non-cataract (SP) | ACC | Non-cataract (SE) | Mild (SE) | Moderate (SE) | Severe (SE) | ACC |
| Ref. [43] | 62.80 | 60.05 | 61.92 | N/A | N/A | N/A | N/A | N/A |
| Ref. [46] | 91.11 | 87.99 | 90.11 | 87.99 | 72.05 | 60.83 | 81.72 | 77.42 |
| Ref. [48] | 93.06 | 92.15 | 92.77 | 92.15 | 75.66 | 74.11 | 95.52 | 83.84 |
| Our method | 95.44 | 93.53 | 94.83 | 93.53 | 82.17 | 73.27 | 89.66 | 85.98 |

Compared to the features of Fourier transform [43], DCT of sketch lines [44–45], luminance and correlation [46], and local standard deviation [48], the detail component of the Haar wavelet indicates the blurriness more directly [47]. Blurring of digital retinal images occurs because no significant gray variation exists between the retinal structures and background [66]. Thus, the intensity difference between neighboring pixels is an immediate method for evaluating the blurriness. The essence of the detail component of the Haar wavelet is simply the intensity difference. Furthermore, the Haar wavelet is improved in this method, and the contrast of one-layer decomposition is enhanced. This may be the reason that our proposed method is superior to the related works and exhibits improved performance over the original Haar wavelet.

Instead of using one threshold value, three sets of two-class classifiers based on different thresholds are employed for the voting. Based on the voting, four non-cataract images, three mild cataract images, five moderate cataract images, and three severe cataract images are correctly classified.

In Table V, the deep learning method [49] exhibits the best performance in both the two-class and four-class classification, although more training samples are used than in other related works. Its performance is significantly dependent on the number of training samples. A large sample size will aid in improving the deep learning method performance, but for small sample size problems, specialized deep learning algorithms should be investigated for effective performance.

The classification results may be affected in retinal images in which the pathology or illumination causes image blurring. Four examples are presented in Figure 21, in which (a) is a pathological myopia retina after laser rectification and (b) illustrates a retinal detachment. Figures 21(c) and (d) are caused by insufficient illumination and light leakage. For retinal images with other pathologies, accurate classification can be obtained if the pathology does not affect the clarity, such as those in Figures 21(a) and (d). Furthermore, Figure 21(b) is a non-cataract case, but it is classified as a mild cataract as half of the retinal structure is lost. Figure 21(c) is a mild cataract, but it is classified as a moderate cataract because the image is dark and the retinal structure is difficult to recognize. These images number 125 in total, in which 43 images are misclassified. Therefore, the number of misclassified images constitutes 3.2% of the database.

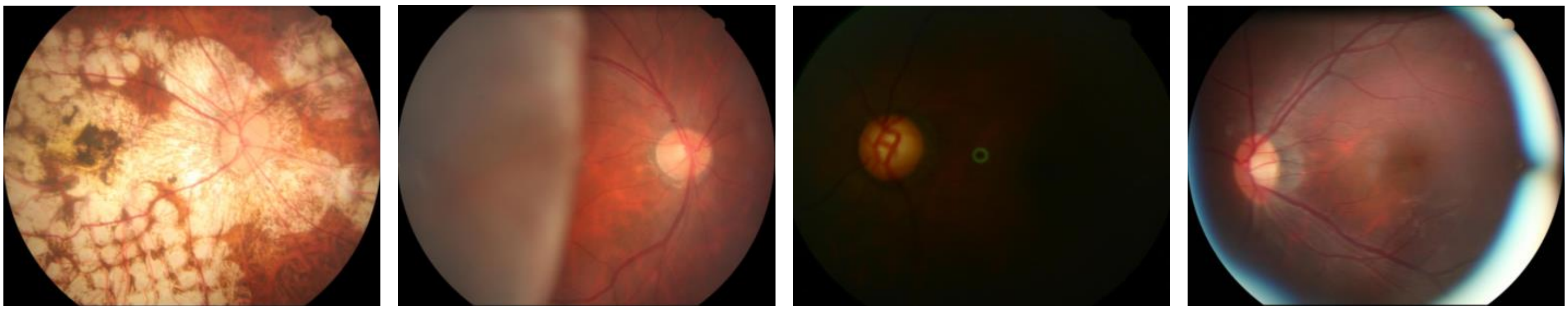

(a) (b) (c) (d)

**Figure 21.** (a) Pathological myopia retina after laser surgery, (b) retinal detachment, (c) insufficient illumination, and (d) light leakage caused by eye rotation.

In this research, the blurriness of retinal images is employed to evaluate the cataract degree. The main cause of image blurring is cataracts, while other eye diseases, such as corneal edema, uveitis, vitreous hemorrhage, and diabetes mellitus, may also lead to blurry retinal images. At present, different types of eye diseases that cause blurring of retinal images cannot be distinguished. Moreover, the three types of age-related cataracts (nuclear cataracts, cortical cataracts, and PSCs) cannot be discriminated. These are the two limitations that should be investigated further.

## 6. Conclusions

An automatic method for cataract detection and grading using retinal images has been proposed. In this method, the improved Haar wavelet feature is used for classification. With the improvement of the Haar wavelet, the contrast of the one-layer detail components is greatly enhanced, and the distortion of the detail components caused by multilayer decomposition is avoided. In the classification of blurry retinal images, two adjacent classes are the most difficult to distinguish. Three sets of two-class classifiers were trained, and the four-class classification problem was transformed into three two-class classification problems. Thereafter, the three sets of two-class classifiers were integrated together to construct a complete classification system. Cross-validation was used to verify the performance. The results demonstrate that the improved method outperforms the original Haar wavelet-based method, and the combination of two-class classifiers is superior to a simple four-class classifier. The analysis indicates that the proposed method can be applied as an add-on function in current fundus photography devices. Moreover, cataract patients may also suffer from other ophthalmological diseases, which may have a negative effect on automatic cataract detection. Thus, future work will focus on the detection of other ophthalmological diseases.

## Acknowledgement

We would like to express our sincere thanks to the Institute of Ophthalmology, Beijing Tongren Hospital, for providing the retinal images.